%% file: main.tex
\documentclass[10pt,twocolumn,letterpaper]{article}

\usepackage{cvpr}              % To produce the CAMERA-READY version

\definecolor{cvprblue}{rgb}{0.21,0.49,0.74}
\usepackage[pagebackref,breaklinks,colorlinks,citecolor=cvprblue]{hyperref}
\usepackage[accsupp]{axessibility} 
\usepackage{color}
\usepackage{colortbl}
\definecolor{graycolor}{rgb}{0.95,0.95,0.95}
\usepackage{multirow}

\usepackage[labelformat=simple]{subcaption}

\usepackage{graphicx}
\usepackage{xfp,graphicx}
\usepackage{amsmath}
\usepackage[accsupp]{axessibility}
\usepackage{natbib} % reference manager

\def\paperID{1275} % *** Enter the Paper ID here
\def\confName{CVPR}
\def\confYear{2025}

\title{Distilling Monocular Foundation Model for Fine-grained Depth Completion}

\author{Yingping Liang\textsuperscript{1} \quad 
Yutao Hu\textsuperscript{2} \quad
Wenqi Shao\textsuperscript{3} \quad
Ying Fu\textsuperscript{1$\dagger$} \\
	\textsuperscript{1}Beijing Institute of Technology \qquad \textsuperscript{2}Key Laboratory of New Generation Artificial Intelligence \\ Technology and Its Interdisciplinary Applications, Southeast University \qquad \textsuperscript{3}Shanghai Al Laboratory\\
	{\tt\small\{liangyingping,fuying\}@bit.edu.cn \quad huyutao@seu.edu.cn \quad weqish@link.cuhk.edu.hk}}

\begin{document}
\twocolumn[{%
\renewcommand\twocolumn[1][]{#1}%
\maketitle

\begin{center}
    \centering
    \includegraphics[width=0.95\textwidth]{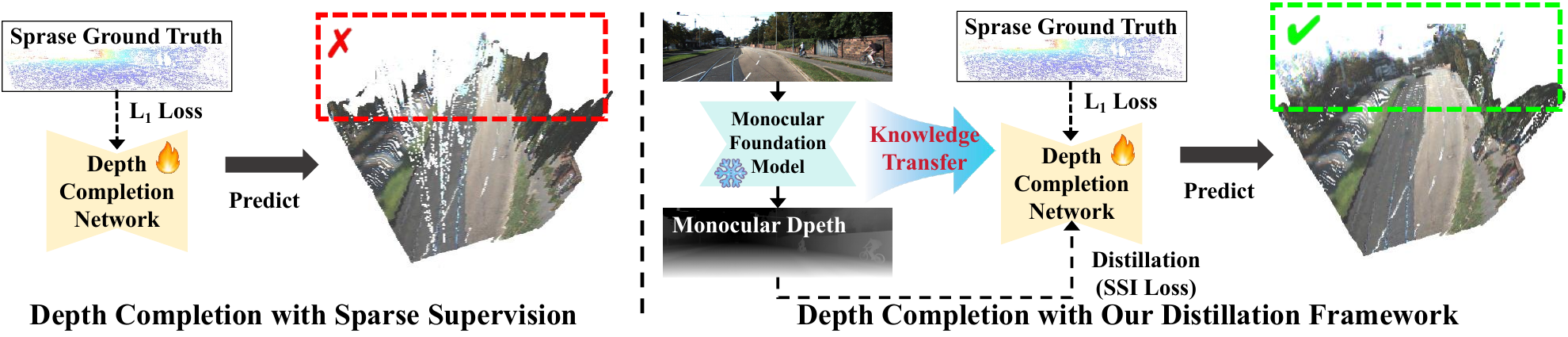}
    \captionof{figure}{Depth completion models trained solely with $L_1$ loss and sparse ground truth produce incomplete and fragmented depth predictions. Our framework, however, demonstrates significant improvements by distilling knowledge from monocular foundation models and incorporating a scale- and shift-invariant loss (SSI Loss), resulting in more complete and accurate dense depth completion.}
    \label{fig:teaser}
\end{center}%
}]

{
\renewcommand{\thefootnote}{}
\footnotetext[5]{$\dagger$ Corresponding author.} 
}

\begin{abstract}
    Depth completion involves predicting dense depth maps from sparse LiDAR inputs. However, sparse depth annotations from sensors limit the availability of dense supervision, which is necessary for learning detailed geometric features. In this paper, we propose a two-stage knowledge distillation framework that leverages powerful monocular foundation models to provide dense supervision for depth completion. In the first stage, we introduce a pre-training strategy that generates diverse training data from natural images, which distills geometric knowledge to depth completion. Specifically, we simulate LiDAR scans by utilizing monocular depth and mesh reconstruction, thereby creating training data without requiring ground-truth depth. Besides, monocular depth estimation suffers from inherent scale ambiguity in real-world settings. To address this, in the second stage, we employ a scale- and shift-invariant loss (SSI Loss) to learn real-world scales when fine-tuning on real-world datasets. Our two-stage distillation framework enables depth completion models to harness the strengths of monocular foundation models. Experimental results demonstrate that models trained with our two-stage distillation framework achieve state-of-the-art performance, ranking \textbf{first place} on the KITTI benchmark. Code is available at \url{https://github.com/Sharpiless/DMD3C}
\end{abstract}

\section{Introduction}
\label{sec:intro}

Depth completion is a fundamental task in computer vision, where the goal is to generate dense depth maps from sparse depth measurements. This task is particularly important in applications such as autonomous driving \cite{jeon2022struct, wu2022sparse, serhatoglu2022rgb}, robotics \cite{khan2022comprehensive, xie2024depth, maffra2019real}, and augmented reality \cite{hu2022deep, li2024supervise}.

Recent approaches \cite{lee2019depth, zhang2018deep, xu2019depth, lee2021depth, chen2019learning, yan2022rignet} leverage deep neural networks to learn from depth data. Some methods \cite{zhang2023completionformer, wang2023lrru} also incorporate RGB images to guide the depth completion process. Despite these advancements, state-of-the-art models \cite{yan2024tri, wang2024improving, tang2024bilateral} still face difficulties in capturing fine-grained geometric details, particularly in complex outdoor scenes where depth annotations are sparse \cite{Uhrig2017THREEDV, wei2021fine, zhang2024atlantis}.

The challenge arises from the reliance on sparse ground truth for training. The lack of dense ground truth makes it difficult for models to accurately learn depth completion across an entire scene. Monocular depth estimation \cite{Ranftl2022, yang2024depth, depth_anything_v2, ke2024repurposing}, on the other hand, has the ability to provide dense depth predictions from a single image. Specifically, state-of-the-art monocular foundation models could generate dense depth maps, containing fine-grained details and relative depth relationships, offering valuable guidance for training depth completion networks.

To make full use of the advantages of monocular foundation models, we propose a novel two-stage distillation framework to transfer geometric knowledge from monocular foundation models to depth completion networks. Specifically, in the \textbf{first distillation stage}, we generate training data through monocular foundation models and then distill knowledge via the proposed pre-training strategy. Specifically, diverse natural images are used to generate pseudo depth maps. Then, we utilize randomly sampled camera parameters to re-construct the scene with mesh and simulate LiDAR using ray simulation. The generated data trains the depth completion model to learn diverse geometric knowledge from monocular foundation models, enhancing its ability to generalize across different scenes.

However, monocular depth estimation suffers from inherent scale ambiguity \cite{zheng2024multi, guan2023hrpose}, resulting in depth predictions that vary greatly in scale. Thus, monocular depth alone cannot serve as a reliable basis for real-world depth. To solve this problem, in the \textbf{second distillation stage}, we introduce a scale- and shift-invariant loss (SSI Loss) \cite{Ranftl2022}. Specifically, when fine-tuning on labeled datasets with sparse ground truth, SSI Loss ignores the scale and shift that causes the least loss from the depth prediction and monocular depth, to ensure consistent depth completions aligning monocular depth supervision across varying scales. 

By integrating these two distillation stages into the training process, our method achieves state-of-the-art performance on the KITTI benchmark. The contributions of this work are summarized as follows:

\begin{itemize}
    \item We propose a novel two-stage distillation framework to transfer knowledge from monocular foundation models to depth completion models, which enables the learning of fine-grained depth information from sparse ground truth by providing dense supervision.
    
    \item In the first stage, we propose a data generation strategy that uses monocular depth estimation and mesh reconstruction to simulate training data, enabling the model to learn geometric features from diverse natural images without the need for any LiDAR and ground truth.

    \item In the second stage, we propose a scale- and shift-invariant loss (SSI Loss) to address the scale ambiguity problem in monocular depth estimation, which ensures consistent depth completions across varying scales and focuses on learning real-world scale information.

    \item Our method achieves first place on the official KITTI depth completion benchmark, demonstrating the effectiveness of the proposed approach with significant gains in both quantitative metrics and qualitative visualizations.
\end{itemize}

\section{Related Work}
\label{sec:related}

\subsection{Depth Completion Datasets}

Depth completion has been widely studied in both indoor and outdoor settings. Indoor datasets, such as NYU Depth V2 \cite{Silberman2012nyu2} and Matterport3D \cite{Chang2017Matterport3D}, are popular because dense depth measurements are easier to obtain in controlled environments. NYU Depth V2, for instance, provides dense depth maps from a Kinect sensor along with RGB images.

In outdoor environments, however, obtaining dense depth annotations is more challenging due to limitations in sensing technologies like LiDAR. The KITTI dataset \cite{Uhrig2017THREEDV} is a key benchmark for outdoor depth completion but suffers from sparse annotations, covering only about 5\% of the image. To address this, training methods often employ complex post-processing and multi-frame fusion techniques to increase the annotation coverage to around 20\% at most. Furthermore, ground truth values for long distances and dynamic objects typically cannot be included. The sparse ground truth raises challenges for training fine-grained depth completion models. Therefore, we employ monocular depth estimation techniques for dense distillation to preserve fine-grained details.

\begin{figure*}[t]
\begin{center}
    \includegraphics[width=0.85\linewidth]{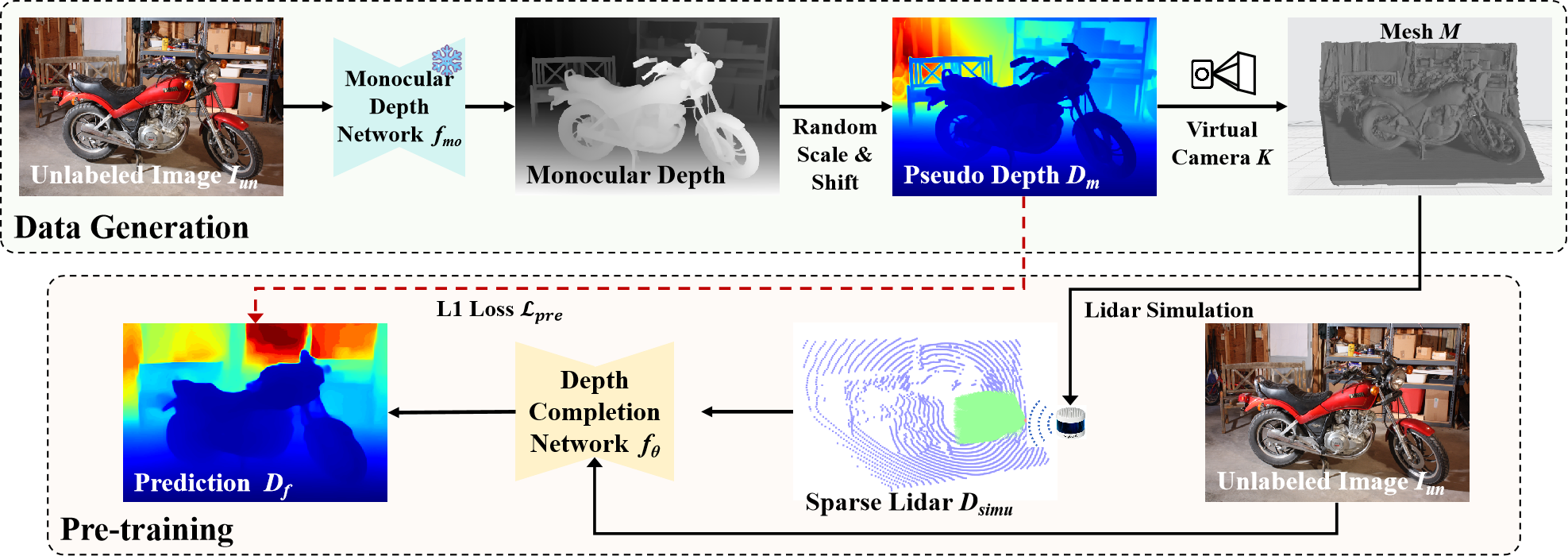}
    \end{center}
    \caption{Illustration of our proposed first distillation stage with a data generation strategy to learn geometric features from monocular foundation models, which only requires unlabeled RGB images. We use the estimated monocular depth to re-construct the scene and then simulate the Lidar swap process to generate sparse points for training.}
    \label{fig:mono2lidar}
\end{figure*}

\subsection{Depth Completion Methods}

Depth completion methods have seen significant advancements \cite{khan2022comprehensive, hu2022deep, 9875218}. Compared to predicting depth directly from a single RGB image \cite{bhat2023zoedepth, piccinelli2024unidepth, yin2023metric3d, 10638254}, depth completion fuses LiDAR information to obtain more accurate sparse depth cues. Recent methods \cite{ma2018sparse, 2024freqfusion, cheng2019learning, qiu2019deeplidar, cheng2020cspn++, tang2020learning, liu2021fcfr, zhao2021adaptive, park2020non, yu2023aggregating, yan2022rignet, lin2022dynamic, zhou2023bev, zhang2023completionformer, wang2023lrru, yan2024tri, tian2023transformer, wang2024improving, zhang2024deep, tang2024bilateral, yan2022multi, yan2023desnet, liu2024transformer, yan2023distortion, conti2023sparsity, liu2024siamese, bartolomei2024revisiting, zou2024ogni, park2024test, wu2025augundo, wang2024scale, park2024simple} rely on supervised learning, using pairs of sparse depth maps and corresponding dense ground truth. While effective with sufficient labeled data, these methods are limited in real-world applications, where obtaining dense annotations, especially in outdoor environments, is challenging \cite{chen2023instance, fu2022low, zou2024eventhdr}. To reduce the reliance on dense labels, \cite{zhang2024self, choi2021stereo} use a teacher-student model, where a teacher network trained on labeled data provides guidance to a student network. However, pseudo-labels in complex or dynamic scenes can be noisy or inaccurate. Therefore, we turn to monocular depth estimation to provide fine-grained supervision for depth completion. 

\subsection{Monocular Depth Estimation}

Monocular depth estimation \cite{Ranftl2022, yang2024depth, depth_anything_v2, ke2024repurposing} has emerged as a powerful tool for generating dense depth maps from single RGB images, making it particularly valuable in scenarios where dense depth annotations are sparse or unavailable \cite{liang2023mpi, guo2024camera, guo2024lidar}. Notable models, such as MiDaS \cite{Ranftl2022} and Depth-Anything \cite{yang2024depth, depth_anything_v2}, have demonstrated impressive capabilities in predicting dense depth maps across a wide range of scenes. G2-MonoDepth \cite{wang2023g2} have proposed a general framework for monocular RGB-X Data with a scale invariant loss. Despite their versatility, monocular depth estimation methods face the challenge of scale ambiguity, which can limit the accuracy of depth completion, especially when integrated with sparse depth measurements from LiDAR. Therefore, we introduce a scale- and shift-invariant loss inspired by \cite{wang2023g2}, combined with a supervised loss to keep real-world scales when fine-tuning on real-world datasets.

\section{Method}
\label{sec:method}

\begin{figure*}[t]
\begin{center}
    \includegraphics[width=0.9\linewidth]{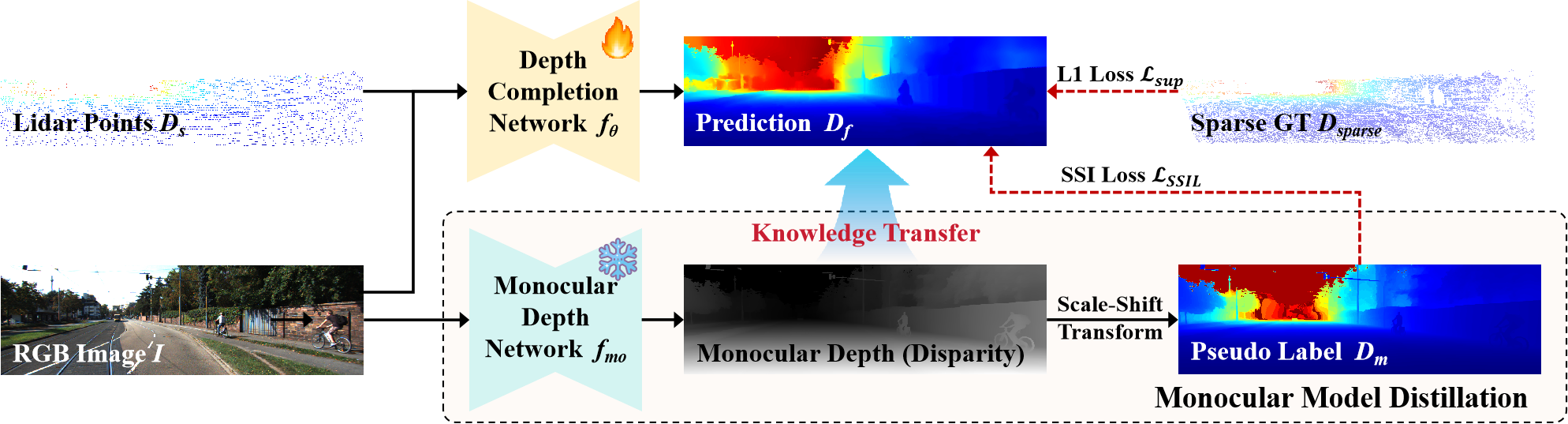}
    \end{center}
    \caption{Illustration of our proposed second distillation stage utilizing foundation models for monocular depth estimation when fine-tuning on labeled datasets. Sparse ground truth provides real-world depth scale with L1 loss. Our method enhances this process by incorporating dense monocular depth for fine-grained supervision. However, monocular depth maps come with inherent scale and shift ambiguities. To address these challenges, we employ a Scale- and Shift-Invariant Loss (SSI Loss) that aligns the predictions with the dense monocular depth to match the real-world depth scale, ensuring more accurate depth completion.}
    \label{fig:framework}
\end{figure*}

In this section, we first provide a brief overview of the motivation and formulation of our method. Then, we introduce our two-stage distillation framework. In the first stage, we propose a data generation strategy for pre-training that leverages large-scale, unlabeled images without ground-truth depth annotations, as illustrated in Figure \ref{fig:mono2lidar}. In the second stage, we detail the fine-tuning process on a labeled dataset with sparse ground truth, using monocular depth in combination with SSI Loss, as shown in Figure \ref{fig:framework}.

\subsection{Motivation and Formulation} 

In the depth completion task, the input typically consists of an RGB image \( I \) and a sparse depth map \( D_s \) obtained from a LiDAR sensor. The objective is to produce a dense depth map \( D_f \) that provides a complete and detailed representation of the scene’s depth. To achieve this, a depth completion model \( f_\theta \), parameterized by \( \theta \), takes both the RGB image and sparse depth map as inputs:
\begin{eqnarray}
    D_f = f_\theta(I, D_s).
\end{eqnarray}

To guide the model’s learning, a sparse supervision loss is defined based on the available sparse depth annotations:
\begin{eqnarray}
\mathcal{L}_{\text{sup}} = M \times \left| D_f - D_{\text{sparse}} \right|,
\label{equ:supervised}
\end{eqnarray}
where \( M \) represents the valid mask for sparse depth ground truth \( D_{\text{sparse}} \). This loss aligns the predicted dense depth map \( D_f \) with the sparse ground truth, ensuring that the output is consistent with the available depth data. However, this sparse supervision provides limited guidance, particularly in outdoor scenes \cite{Uhrig2017THREEDV}, leading to challenges in achieving fine-grained and consistent depth predictions due to the inherent limitations of sparse supervision. 

Foundation models for monocular depth estimation \cite{yang2024depth, depth_anything_v2} provide an alternative approach by generating fine-grained dense depth from single RGB images, which we use as supervision without relying solely on LiDAR. Therefore, in the first stage, we utilize the monocular foundation models to generate diverse and large-scale training data, to provide dense supervision for pre-training, allowing the model to learn geometric features. However, monocular methods suffer from scale ambiguities, which can lead to inconsistencies when combining dense monocular predictions with sparse depth. To address this, in the second stage, we introduce SSI Loss \cite{Ranftl2022} combined with L1 loss when fine-tuning on labeled datasets with sparse ground truth to mitigate scale inconsistencies and learn real-world scale. 

\subsection{First Stage: Data Generation and Pre-training}
\label{sec:pretrain}

To leverage diverse images without depth annotations, we introduce a pre-training strategy that utilizes synthesized depth data generated through monocular depth estimation on large-scale natural image datasets inspired by \cite{2024relation, liang2023mpi}. As illustrated in Figure \ref{fig:mono2lidar}, this pre-training phase enables the model to learn robust geometric features from monocular foundation models across diverse scenes, preparing it for subsequent fine-tuning on real-world datasets.

Specifically, we first utilize a pre-trained monocular depth estimation model, such as Depth Anything V2 \cite{depth_anything_v2}, to predict depth maps for such images. For each natural image \( I_{\text{un}} \), the monocular depth estimation model predicts a corresponding dense depth map \( D_{\text{syn}} \):
\begin{eqnarray}
    D_{\text{syn}} = f_{\text{mo}}(I_{\text{un}}),
\end{eqnarray}
where \( f_{\text{mo}} \) represents the monocular depth estimation model. These synthesized depth maps, though not accurate in metric scale, capture the relative depth relationships and structural details in the scene, providing valuable supervision signals during pre-training. 

To simulate LiDAR scanning, we first sample a random camera intrinsic matrix \( K \) as:
\begin{eqnarray}
K = \begin{bmatrix} f_x & 0 & c_x \\ 0 & f_y & c_y \\ 0 & 0 & 1 \end{bmatrix},
\end{eqnarray}
where \( f_x \) and \( f_y \) represent the focal lengths along the \( x \)- and \( y \)-axes, and \( (c_x, c_y) \) denotes the center of the image.

For each pixel \( (u, v) \) in the depth map, the corresponding 3D coordinates \( (X, Y, Z) \) in the camera coordinate system are derived using the sampled camera intrinsic matrix \( K \). This transformation generates a point cloud \( P = \{(X, Y, Z)\} \) that represents the 3D spatial locations of each pixel. Leveraging the 3D point cloud \( P \), we reconstruct a mesh \( M \) via surface reconstruction techniques, such as Poisson Surface Reconstruction \cite{kazhdan2006poisson}, to create a continuous 3D surface model for simulation.

Next, we simulate a LiDAR sensor by generating a set of ray direction vectors \( \mathbf{d} \). Each ray is cast from the origin along \( \mathbf{d}_{i,j} \), intersecting the mesh \( M \) to determine the distance to the intersection point. This distance is recorded as the simulated LiDAR depth reading \( D_{\text{simu}}(i, j) \). The resulting sparse depth map \( D_{\text{simu}} \) emulates LiDAR depth readings from the reconstructed 3D surface, to provide a synthetic depth image comparable to actual LiDAR data.

Then, the depth completion model $f_\theta$ takes the RGB image and the sparse depth from simulated scanning as inputs:
\begin{eqnarray}
    D_f = f_\theta(I_{\text{un}}, D_{\text{simu}}),
\end{eqnarray}
where $D_{\text{simu}}$ is the sparse input from simulated LiDAR scanning. Afterwards, the depth completion model is pre-trained using L1 Loss:
\begin{align}
    % SSIL(D_f, & D_m) = \\ 
    \mathcal{L}_{\text{pre}} = \left| D_f - D_{m} \right|.
\end{align}

This pre-training process encourages the depth completion model \( f_\theta \) to align its predictions with the synthesized depth maps, learning meaningful geometric features from the generated training data using monocular foundation models and diverse natural image data. Additionally, the model learns complex geometric structures and relative depth relationships across varying scenes, which are crucial for accurate depth completion in the fine-tuning phase.

\subsection{Second Stage: Fine-tuning} 
\label{sec:distillation}

The depth completion model pre-trained on the generated data learns powerful geometric features from the monocular foundation models. However, due to the inherent scale ambiguity of monocular depth, the model prediction is not consistent with the scale in the real world. To learn the real-world scale, we introduce a combined loss for fine-tuning on labeled datasets. To be specific, we first utilize the sparse ground truth \( D_s \) for supervised loss $\mathcal{L}_{\text{sup}}$ to focus on real-world depth, as in Equation \ref{equ:supervised}. Then, the dense monocular depth \( D_m \) predicted by monocular depth models is also used to provide approximate depth values across the image, especially in regions not covered by the ground truth \( D_s \). 

However, the key challenge in using monocular depth estimation as supervision is the inherent scale ambiguity in depth predictions. To address this issue, we incorporate a Scale- and Shift-Invariant Loss (SSI Loss) into our distillation framework. The SSI Loss is designed to align the predicted depth map \( D_f \) with the dense monocular depth \( D_m \). It remains invariant to differences in scale and shift between \( D_f \) and \( D_m \). The SSI Loss is formulated as:
\begin{align}
    % SSIL(D_f, & D_m) = \\ 
    \mathcal{L}_{\text{SSIL}} = \min_{s, b} \left| D_f - (s \cdot D_m + b) \right|,
    \label{eq:ssil}
\end{align}
where \( s \) and \( b \) are the optimal scale and shift parameters that align the predicted depth \( D_f \) with the dense monocular depth \( D_m \). The loss function seeks to find the best alignment between \( D_f \) and \( D_m \), effectively normalizing any global differences in scale and offset. By minimizing this loss across all pixels in the image, the model is encouraged to produce depth maps that maintain consistency with the relative depth structure provided by the monocular depth estimates. In addition, we adapt a gradient matching term to preserve sharpness and align with with depth discontinuities. The gradient matching term is defined as:
\begin{align}
\mathcal{L}_{\text{reg}} = \frac{1}{N} \sum_{k=1}^{K} \left( |\nabla_x R^k| + |\nabla_y R^k| \right),
\end{align}
where \( R = D_f - D_m \), and \( R^k \) denotes the difference of depth maps at scale \( k \). We use \( K = 4 \) scale levels as in \cite{Ranftl2022}, halving the image resolution at each level. The final objective function combines the supervised loss \( \mathcal{L}_{\text{sup}} \), the dense distillation loss \( \mathcal{L}_{\text{SSIL}} \), and the regularization term \( \mathcal{L}_{\text{reg}} \).

\begin{table}[t] 
% %  \small
\centering
\caption{Unlabeled datasets used for pre-training data generation.}
\begin{tabular}{@{}lccc@{}}
\toprule
Dataset & Indoor & Outdoor & \# Images \\ \midrule
COCO \cite{lin2014microsoft} & \checkmark & \checkmark & 118,287 \\
Google Landmarks \cite{weyand2020google} & & \checkmark & 117,576 \\
Nuscenes \cite{caesar2020nuscenes} & & \checkmark & 93,475 \\
Cityscapes \cite{cordts2016cityscapes} & & \checkmark & 19,998 \\ 
DAVIS \cite{perazzi2016benchmark} & \checkmark & \checkmark & 10,581 \\
\bottomrule
\end{tabular}
\label{tab:datasets}
\end{table}

\begin{table*}[ht]
\small
\centering
\caption{Performance on KITTI and NYUv2 datasets. For the KITTI dataset, results are evaluated by the KITTI testing server and ranked by the RMSE (in mm). For the NYUv2 dataset, we report their performance on the official test set in their papers.}
\label{tab:performance}
\begin{tabular}{@{}lcccccccc@{}}
\toprule
\multirow{2}{*}{Method} & \multirow{2}{*}{Year} & \multicolumn{4}{c}{KITTI} & \multicolumn{3}{c}{NYUv2} \\
\cmidrule(lr){3-6} \cmidrule(lr){7-9}
 & & \textbf{RMSE} $\downarrow$ & MAE $\downarrow$ & iRMSE $\downarrow$ & iMAE $\downarrow$ & RMSE $\downarrow$ & REL $\downarrow$ & $\delta_{1.25} \uparrow$ \\
\midrule
S2D \cite{ma2018sparse} & 2018 & 814.73 & 249.95 & 2.80 & 1.21 & 0.230 & 0.044 & 97.1 \\
CSPN \cite{cheng2019learning} & 2019 & 1019.64 & 279.46 & 2.93 & 1.15 & 0.117 & 0.016 & 99.2 \\
DeepLiDAR \cite{qiu2019deeplidar} & 2019 & 758.38 & 226.50 & 2.56 & 1.15 & 0.115 & 0.022 & 99.3 \\
CSPN++ \cite{cheng2020cspn++} & 2020 & 743.69 & 209.28 & 2.07 & 0.90 & 0.101 & 0.015 & 99.5 \\
GuideNet \cite{tang2020learning} & 2020 & 736.24 & 218.83 & 2.25 & 0.99 & 0.115 & -- & -- \\
FCFR \cite{liu2021fcfr} & 2021 & 735.81 & 217.15 & 2.20 & 0.98 & 0.106 & 0.015 & 99.5 \\
ACMNet \cite{zhao2021adaptive} & 2021 & 744.91 & 206.09 & 2.08 & 0.90 & 0.105 & 0.015 & 99.4 \\
NLSPN \cite{park2020non} & 2020 & 741.68 & 199.59 & 1.99 & 0.84 & 0.092 & 0.012 & 99.6 \\
PointDC \cite{yu2023aggregating} & 2023 & 736.07 & 201.87 & 1.97 & 0.87 & 0.089 & 0.012 & 99.6 \\
RigNet \cite{yan2022rignet} & 2022 & 712.66 & 203.25 & 2.08 & 0.90 & 0.090 & 0.013 & 99.6 \\
DySPN \cite{lin2022dynamic} & 2022 & 709.12 & 192.71 & 1.88 & 0.82 & 0.090 & 0.012 & 99.6 \\
BEV@DC \cite{zhou2023bev} & 2023 & 697.44 & 189.44 & 1.83 & 0.82 & 0.089 & 0.012 & 99.6 \\
CFormer \cite{zhang2023completionformer} & 2023 & 708.87 & 203.45 & 2.01 & 0.88 & 0.090 & 0.012 & -- \\
LRRU \cite{wang2023lrru} & 2023 & 696.51 & 189.96 & 1.87 & \textbf{0.81} & 0.091 & 0.011 & 99.6 \\
TPVD \cite{yan2024tri} & 2024 & 693.97 & 188.60 & \textbf{1.82} & \textbf{0.81} & 0.086 & \textbf{0.010} & \textbf{99.7} \\
ImprovingDC \cite{wang2024improving} & 2024 & 686.46 & \textbf{187.95} & 1.83 & \textbf{0.81} & 0.091 & 0.011 & 99.6 \\
BP-Net \cite{tang2024bilateral} & 2024 & 684.90 & 194.69 & \textbf{1.82} & 0.84 & 0.089 & 0.012 & 99.6 \\ \midrule
\rowcolor{graycolor} \textbf{DMD$^{3}$C (Ours)} & - & \textbf{678.12} & 194.46 & \textbf{1.82} & 0.85 & \textbf{0.085} & 0.011 & \textbf{99.7} \\
\bottomrule
\end{tabular}
\label{tab:comparison}
\end{table*}

\section{Experiments}

In this section, we first introduce the details of our implementation, as well as the datasets and evaluation metrics for experiments. Then, detailed comparisons are conducted with the state-of-the-art methods. Finally, ablations and discussions are performed to confirm the effectiveness of our proposed main components. Additional analysis is provided in the supplementary materials, along with videos.

\subsection{Experimental Pipeline}

\noindent\textbf{First stage} involves utilizing our proposed data generation strategy to train the depth completion model from scratch, as described in Section \ref{sec:pretrain}. In this stage, we generate training data using the RGB images in Table \ref{tab:datasets} and train the depth completion model using SSI Loss, which does not require any ground truth depth. This stage allows the pre-trained model to focus on learning diverse geometric features without being constrained by labeled data, enhancing generalization across different scenes.

\noindent\textbf{Second stage} involves adapting the pre-trained model on labeled datasets, using supervised loss (L1 loss) with depth annotations and our proposed SSI Loss for knowledge distillation, as described in Section \ref{sec:distillation}. Using L1 loss with sparse ground truth allows the depth completion model to adapt to the real-world depth scale under sparse supervision. Specifically, our proposed monocular model distillation uses SSI Loss to help the model maintain fine-grained detail with dense supervision by distilling monocular foundation models. Fine-tuning and validation are performed on the KITTI and NYU Depth V2 datasets, respectively.

\subsection{Implementation Details}

We follow the setup of recent work \cite{tang2024bilateral, zhang2023completionformer, wang2024improving} and train our model on 4 NVIDIA RTX A100 GPUs. For the monocular foundation model used in distillation, we adopt Depth Anything V2 \cite{depth_anything_v2} due to its robust performance. For the depth completion network architecture, we use BP-Net \cite{tang2024bilateral} as our base model, which is primarily composed of ResNet \cite{he2016deep} blocks. In the first stage, we utilize a mixed dataset collected from a wide range of resources, as shown in Table \ref{tab:datasets}. We select five large-scale public datasets as our unlabeled sources for their diverse scenes, totaling approximately $360,000$ images with varying scenes and image scales. We adopt AdamW with a weight decay of $0.05$ as the optimizer and clip gradients if their norms exceeding $0.1$. In the first stage, our model is pre-trained from scratch for $600$K iterations. In the second stage, the pre-trained model is fine-tuned on the labeled datasets for $300$K iterations using a OneCycle learning rate policy. We set the batch size to 16. The final model is obtained by applying Exponential Moving Average (EMA) with a decay of $0.9999$, which helps stabilize the model parameters during training.

\subsection{Datasets and Evaluation Metrics}

\noindent\textbf{KITTI Depth Completion Dataset \cite{Uhrig2017THREEDV}} serves as a standard benchmark for depth completion in autonomous driving scenarios. It provides over 93,000 training samples, including sparse LiDAR depth maps, their corresponding RGB images, and ground truth depth maps. The sparse LiDAR depth maps are generated from Velodyne LiDAR scans, which typically cover only a small percentage of the image pixels (approximately 5\%). The ground truth depth maps are obtained via multi-frame matching and cover approximately 20\% image pixels. The dataset also includes 1,000 test samples on the private online benchmark. 

\noindent\textbf{NYU Depth V2 Dataset \cite{Silberman2012nyu2}} is also a standard benchmark for indoor scene understanding, consisting of RGB-D data captured with a Kinect sensor. It provides 1,449 densely labeled images from various indoor scenes. Missing ground truth is mainly due to the difference in perspective between the depth sensor and the camera. 

\noindent\textbf{Evaluation Metrics} used for KITTI are root mean squared error (RMSE), mean absolute error (MAE), root mean squared error of the inverse depth (iRMSE), and mean absolute error of the inverse depth (iMAE), among which RMSE is used for ranking. These metrics are obtained by submission and evaluation from a non-public test set. For NYU, we follow the previous work and report the RMSE, mean absolute relative error (REL), and $\delta_{\theta}$, which represents the percentage of pixels whose error is less than a threshold $\theta$.

\begin{figure*}[t]
    \centering
    \subfloat[Frame]{\includegraphics[width=1.7in]{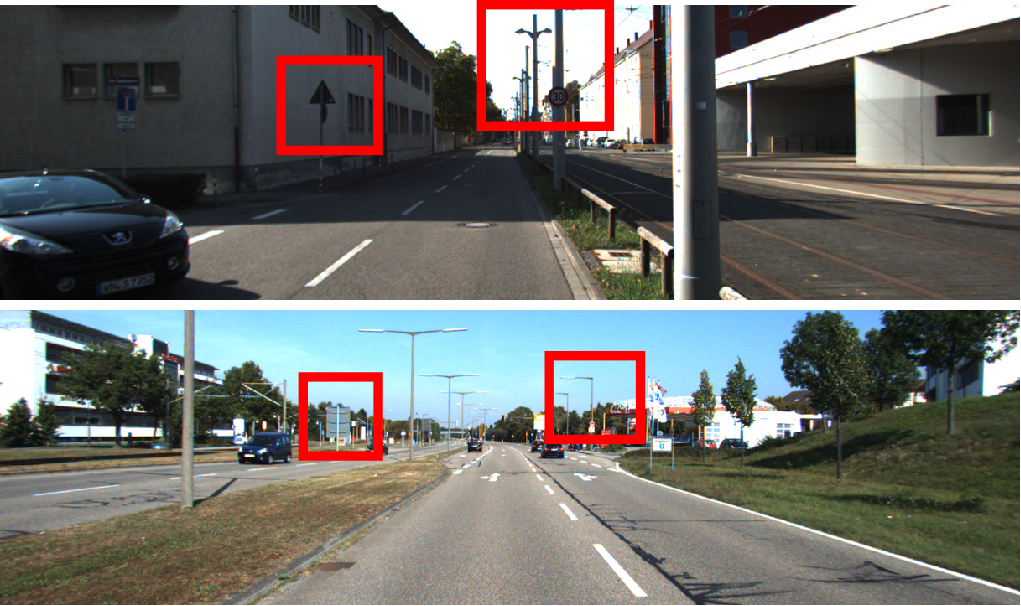}}
    \hfil
    \subfloat[CFormer \cite{zhang2023completionformer}]{\includegraphics[width=1.7in]{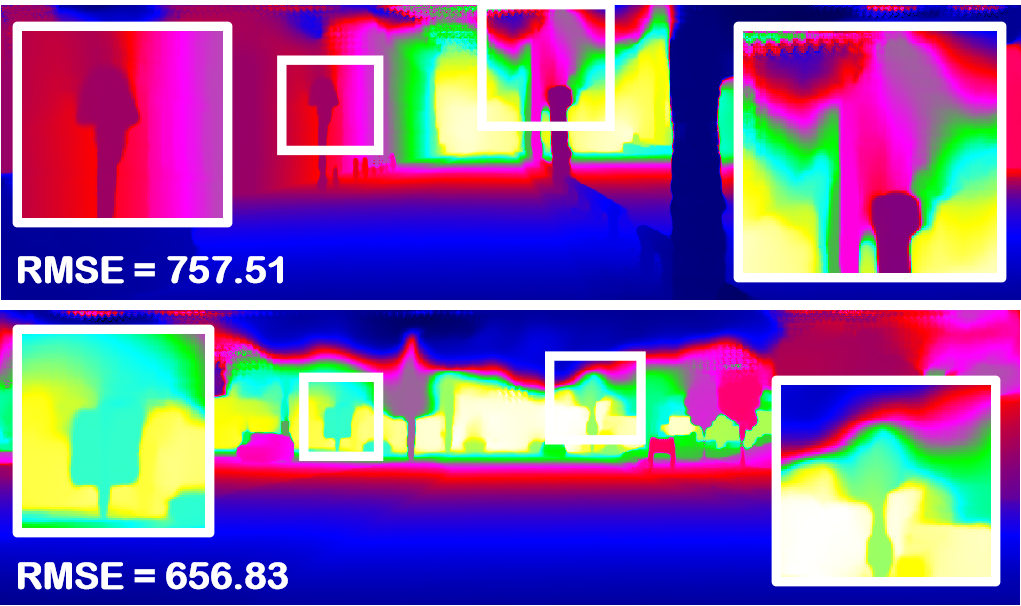}}
    \hfil
    \subfloat[TPVD \cite{yan2024tri}]{\includegraphics[width=1.7in]{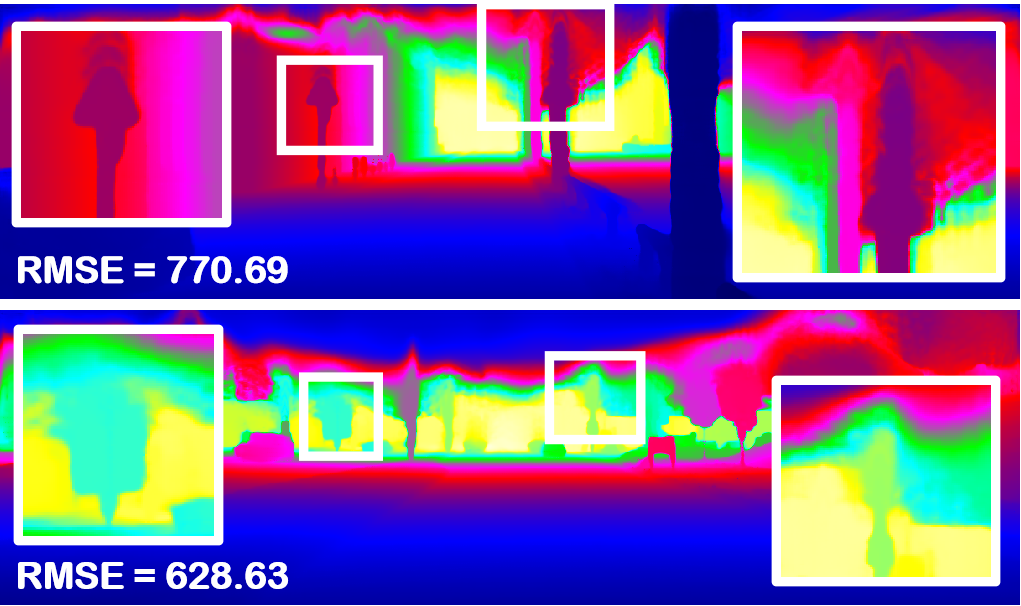}}
    \hfil
    \subfloat[BP-Net \cite{tang2024bilateral}]{\includegraphics[width=1.7in]{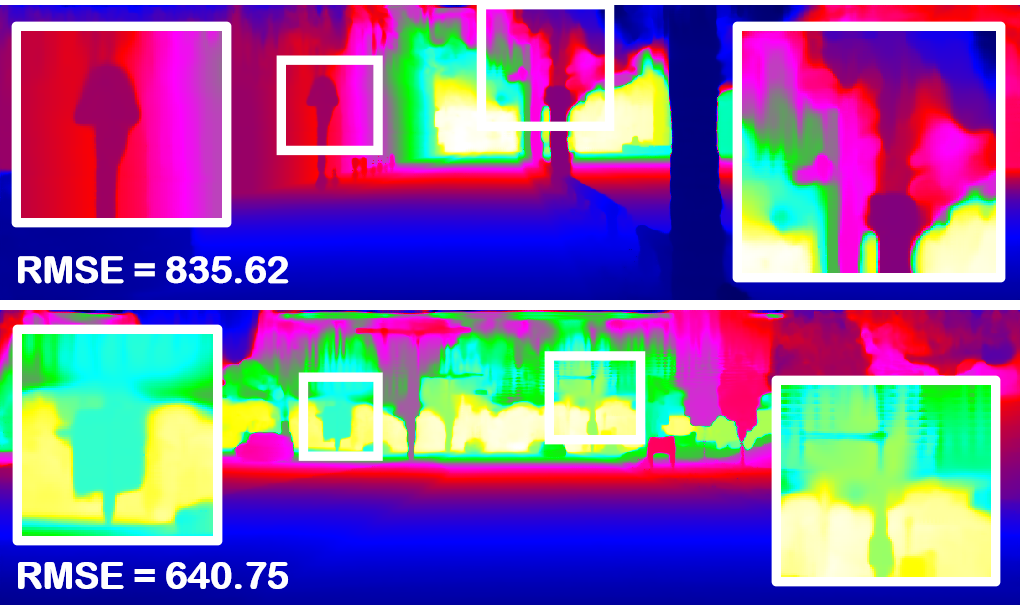}}
    \hfil
    % ########################################################
    \subfloat[Error Map]{\includegraphics[width=1.7in]{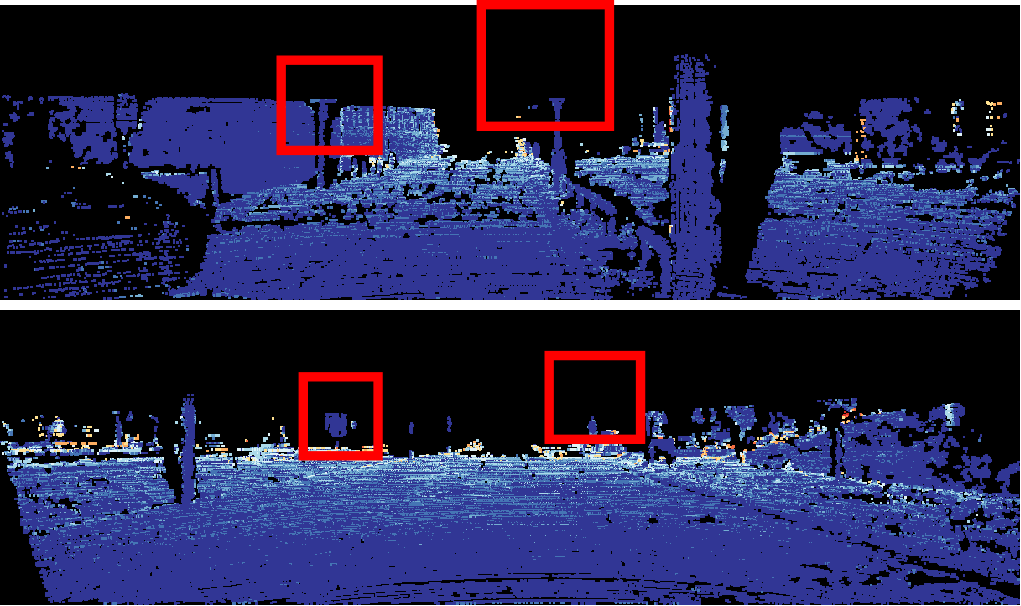}}
    \hfil
    \subfloat[LRRU \cite{wang2023lrru}]{\includegraphics[width=1.7in]{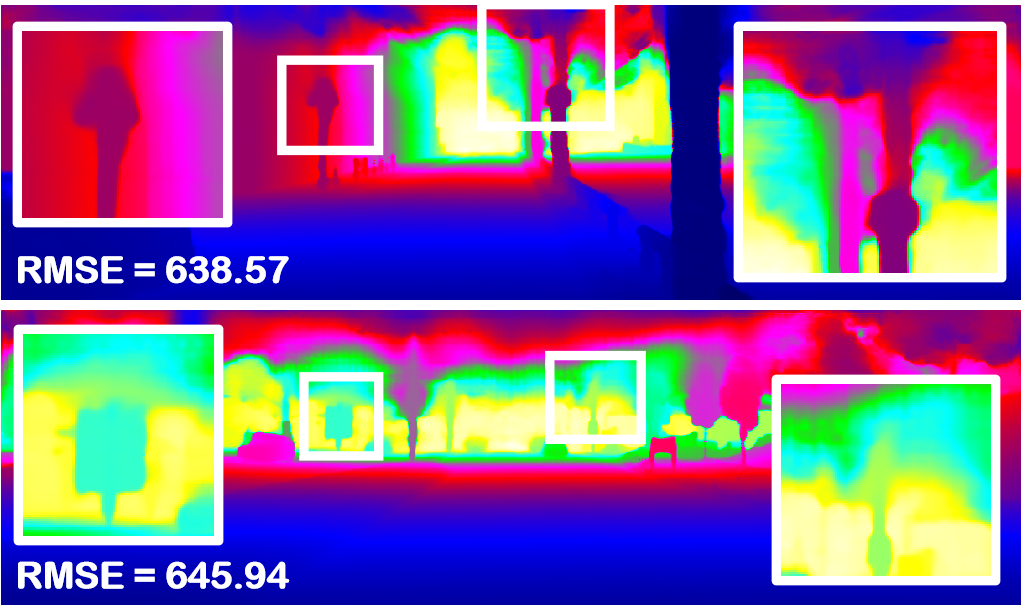}}
    \hfil
    \subfloat[ImprovingDC \cite{wang2024improving}]{\includegraphics[width=1.7in]{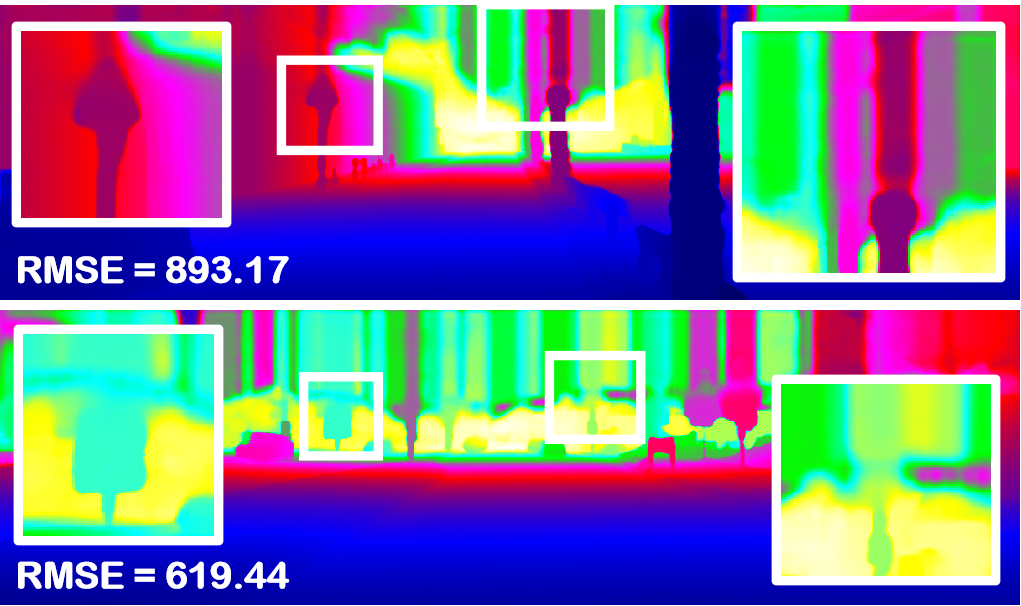}}
    \hfil
    \subfloat[DMD$^{3}$C (Ours)]{\includegraphics[width=1.7in]{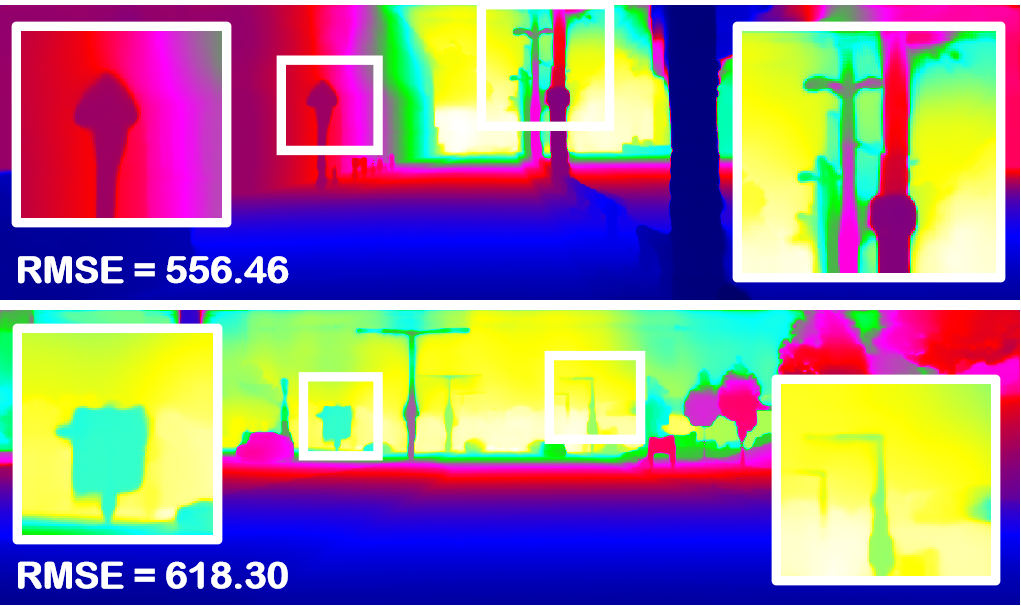}}
    \caption{Qualitative comparison of our proposed DMD$^{3}$C with several state-of-the-art methods on the KITTI benchmark, using public test results. Error maps highlight pixels with ground truth. In regions lacking ground truth, our method demonstrates notable improvements in depth completion, even though these areas are excluded from the evaluation metrics.}
    \label{fig:visualization1}
\end{figure*}

\subsection{Comparison with State-of-the-art Methods}

\noindent\textbf{Evaluation on Depth Completion Benchmarks.} We evaluate our method on the test sets of the NYUv2 dataset \cite{Silberman2012nyu2} and KITTI dataset \cite{Uhrig2017THREEDV}. Table \ref{tab:comparison} shows the quantitative comparison of our method and other top ranking published methods. On the KITTI leaderboard, our proposed DMD$^{3}$C ranks 1st outperforming all other methods under the primary RMSE metric at the time of paper submission. It also has comparable performance under other evaluation metrics. On the NYUv2 dataset, our method achieves the best RMSE and best $\delta_{1.25}$. Additionally, DMD$^{3}$C maintains competitive results across several other evaluation metrics, further confirming its robustness and generalization.

To enhance understanding of DMD$^{3}$C's superiority over other state-of-the-art methods, we provide visual comparisons. Figure \ref{fig:visualization1} provides qualitative comparisons with other SOTA methods on the KITTI test set using publicly available results. We use results from the official benchmark\footnote{\url{https://www.cvlibs.net/datasets/kitti/}} produced by the best models for method comparison. Our DMD$^{3}$C model excels in maintaining sharp object boundaries and capturing fine details in areas where other models encounter difficulties, particularly in scenes with complex structures or objects at varying distances. In such challenging regions, other methods fail to estimate accurate depth. Furthermore, as shown in the error maps, our method demonstrates a clear advantage even in regions without ground truth, despite these areas not being factored into the evaluation metrics.

\begin{figure*}[t]
    \centering
    \subfloat[Input]{\includegraphics[width=1.1in]{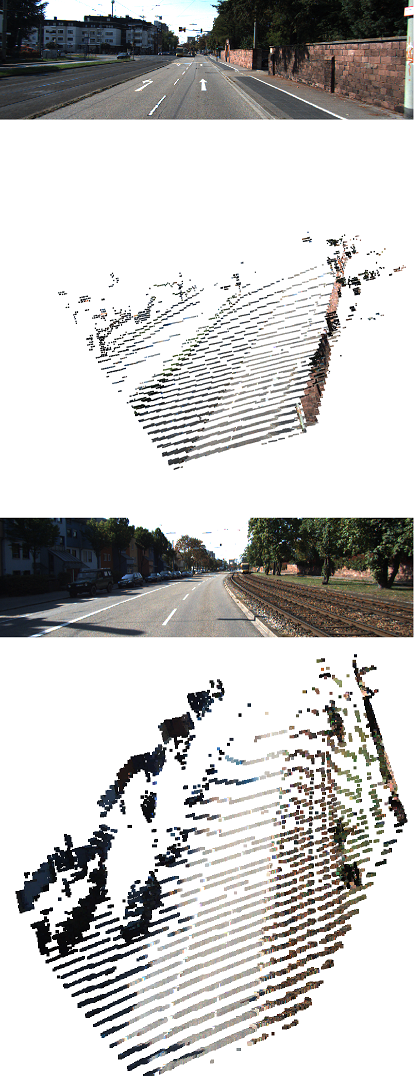}}
    \hfil
    \subfloat[CFormer \cite{zhang2023completionformer}]{\includegraphics[width=1.1in]{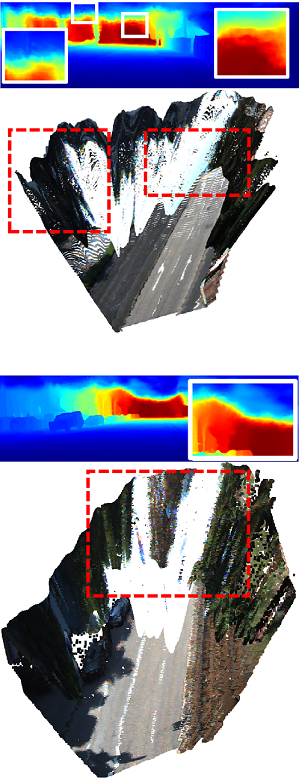}}
    \hfil
    \subfloat[LRRU \cite{wang2023lrru}]{\includegraphics[width=1.1in]{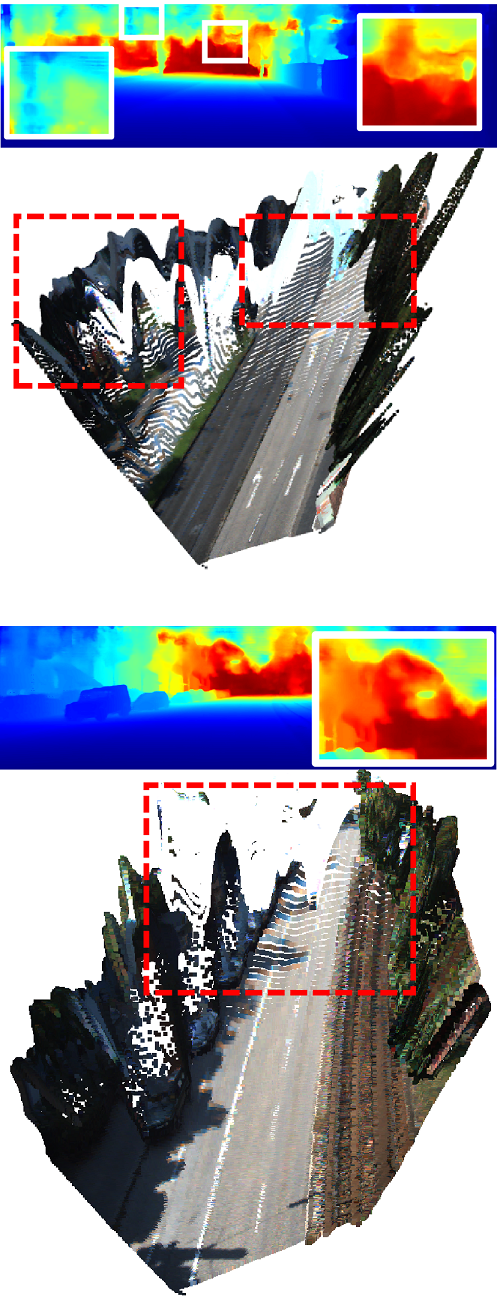}}
    \hfil
    \subfloat[ImprovingDC \cite{wang2024improving}]{\includegraphics[width=1.1in]{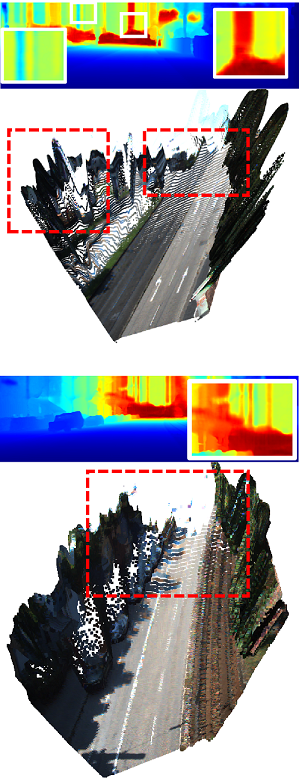}}
    \hfil
    \subfloat[BP-Net \cite{tang2024bilateral}]{\includegraphics[width=1.1in]{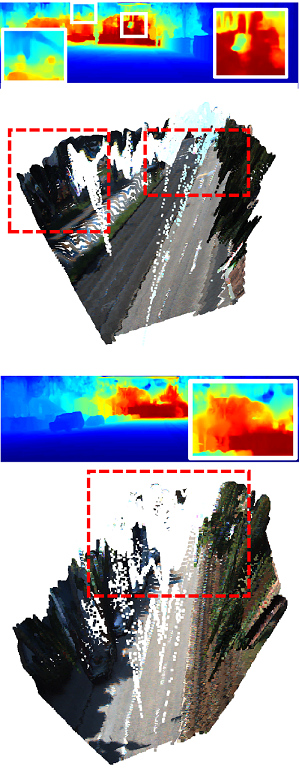}}
    \hfil
    \subfloat[DMD$^{3}$C (Ours)]{\includegraphics[width=1.1in]{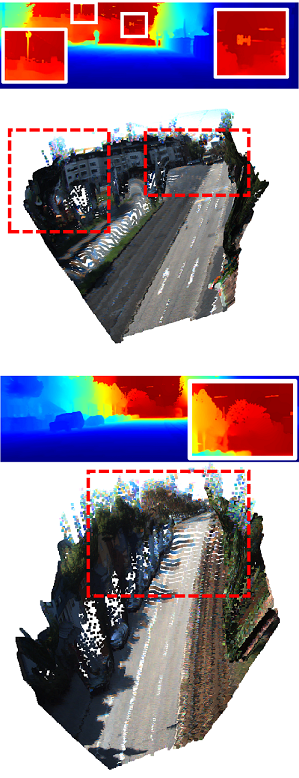}}
    \caption{Qualitative comparison of depth completion methods. This figure demonstrates the performance of various depth completion models, including CFormer, LRRU, ImprovingDC, BP-Net, and our proposed DMD$^{3}$C. For each method, we show the input RGB images with sparse LiDAR points (left), along with the resulting completed depth maps and corresponding 3D point cloud reconstructions.}
    \label{fig:points}
\end{figure*}

Not only does our model retain the fine details in the depth map, but it also maintains structural integrity in the potential three-dimensional space. Figure \ref{fig:points} illustrates that our proposed two-stage distillation improves performance in the 3D space by visualizing point clouds from the completed depth maps. In comparison to other methods, our DMD$^{3}$C produces a more coherent and complete 3D structure, further outperforming existing methods.

\begin{table}[t]
\small
\centering
\caption{Ablations on the proposed main components.}
\label{tab:ablations}
\begin{tabular}{lcccc}
\toprule
\multirow{2}{*}{Method} & \multicolumn{4}{c}{KITTI} \\ \cmidrule(lr){2-5}
~ & RMSE $\downarrow$ & MAE $\downarrow$ & iRMSE $\downarrow$ & iMAE $\downarrow$ \\
\midrule
w/o Pre-train & 682.34 & 194.96 & \textbf{1.82} & \textbf{0.84} \\
w/o SSI Loss & 684.54 & 195.65 & 1.86 & 0.85 \\
\rowcolor{graycolor} DMD$^{3}$C & \textbf{678.12} & \textbf{194.46} & \textbf{1.82} & 0.85 \\
\bottomrule
\end{tabular}
\end{table}

\subsection{Discussions} 

\begin{table}[t]
\small
\centering
\caption{Ablations on different network architectures.}
\label{tab:networks}
\begin{tabular}{lcccc}
\toprule
\multirow{2}{*}{Method} & \multicolumn{4}{c}{KITTI} \\ \cmidrule(lr){2-5}
~ & RMSE $\downarrow$ & MAE $\downarrow$ & iRMSE $\downarrow$ & iMAE $\downarrow$ \\
\midrule
 LRRU \cite{wang2023lrru} & 696.51 & 189.96 & 1.87 & 0.81 \\
\rowcolor{graycolor}\  + Ours & \textbf{693.17} & \textbf{189.60} & \textbf{1.85} & \textbf{0.80} \\
\midrule
 CFormer \cite{zhang2023completionformer} & 764.87 & 183.88 & 1.89 & \textbf{0.80} \\ 
\rowcolor{graycolor}\  + Ours & \textbf{760.29} & \textbf{183.62} & \textbf{1.88} & \textbf{0.80} \\
\midrule
 BP-Net \cite{tang2024bilateral} & 684.90 &  194.69 & \textbf{1.82} & \textbf{0.84} \\
\rowcolor{graycolor}\  + Ours & \textbf{678.12} & \textbf{194.46} & \textbf{1.82} & 0.85 \\
\bottomrule
\end{tabular}
\end{table}

\noindent\textbf{Effect of Pre-training in the First Stage.} First, to evaluate the impact of the proposed pre-training strategy using unlabeled images, we conduct an experiment where the depth completion model is trained from scratch, without any pre-training. As shown in Table \ref{tab:ablations}, removing the pre-training step results in a performance degradation, with an increase in RMSE by $4.22$ mm. This demonstrates that pre-training plays a crucial role in enhancing the model’s ability to generalize, enabling it to better capture complex geometric features across diverse scenes.

\noindent\textbf{Effect of SSI Loss in the Second Stage.} Next, we show the impact of removing SSI Loss in the second distillation stage, while retaining the standard supervised loss with sparse ground truth. As shown in Table \ref{tab:ablations}, the results indicate that the proposed SSI Loss improves the model's performance. This demonstrates the benefit of aligning monocular depth supervision with real-world scale using SSI Loss.

\noindent\textbf{Network Architectures.} We evaluate the compatibility of our method with different network architectures, as shown in Table \ref{tab:networks}. Since we focus on training strategies, we have the flexibility to various models, demonstrating the robustness of our approach across different network designs. Specifically, we evaluate BP-Net \cite{tang2024bilateral}, LRRU \cite{wang2023lrru}, and CFormer (L1 Loss) \cite{zhang2023completionformer}. BP-Net is one of the most advanced methods, while LRRU and CFormer are considered representative architectures. Our approach consistently improves performance across all three models. Notably, BP-Net with our method achieves the best performance in terms of RMSE, the primary evaluation metric. 

\begin{table}[t]
\scriptsize
\centering
\caption{Zero-shot performance comparison with other methods on out-of-the-domain datasets.}
\label{tab:performance}
\begin{tabular}{@{}lcccccc@{}}
\hline
\multirow{2}{*}{Method} & \multicolumn{2}{c}{ScanNet} & \multicolumn{2}{c}{DDAD} & \multicolumn{2}{c}{VOID1500} \\
 & RMSE $\downarrow$ & MAE $\downarrow$ & RMSE $\downarrow$ & MAE $\downarrow$ & RMSE $\downarrow$ & MAE $\downarrow$ \\
\hline
CFormer & 0.120 & 0.232 & 9.606 & 3.328 & 0.726 & 0.261 \\
LRRU & 0.132 & 0.245 & 9.164 & 2.738 & 0.698 & 0.232 \\
BP-Net & 0.122 & 0.212 & 8.903 & 
2.712 & 0.704 & 0.230 \\ 
\textbf{Ours} & \textbf{0.101} & \textbf{0.210} & \textbf{7.766} & \textbf{2.498} & \textbf{0.676} & \textbf{0.225} \\ \hline
\end{tabular}
\label{tab1}
\end{table}

\noindent\textbf{Generalization to Out-of-Domain Datasets.} To evaluate the generalization ability of our method, we conduct zero-shot testing on unseen datasets, including ScanNet \cite{dai2017scannet}, DDAD \cite{guizilini20203d}, and VOID1500 \cite{wong2020unsupervised}, as shown in Table \ref{tab1}. Our approach consistently achieves the best performance across all datasets, demonstrating lower RMSE and MAE compared to prior methods. Notably, our method outperforms BP-Net by a significant margin on DDAD, reducing RMSE from 8.903 to 7.766, highlighting its effectiveness in handling unseen scenarios.

\begin{figure}[t]
\begin{center}
    \includegraphics[width=0.9\linewidth]{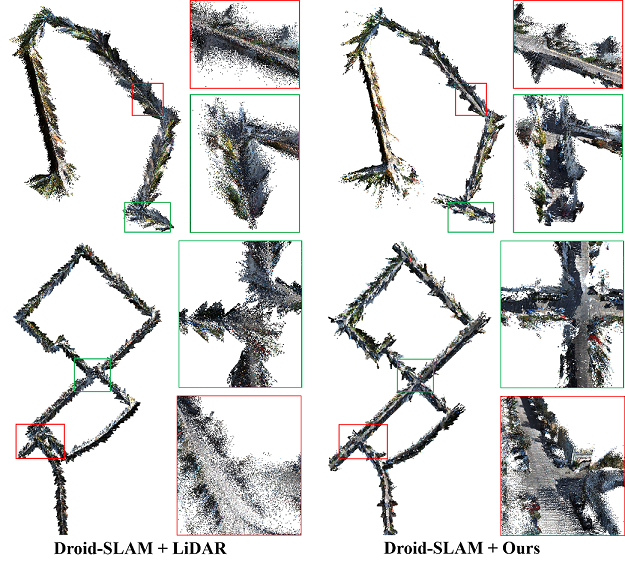}
    \end{center}
    \caption{Qualitative comparison of Droid-SLAM with sparse LiDAR points and our method with dense depth completion.}
    \label{fig:slam}
\end{figure}

\noindent\textbf{Application on Dense SLAM.}
Figure \ref{fig:slam} compares a representative SLAM method, Droid-SLAM \cite{teed2021droid}, with sparse LiDAR points (left) and our method with dense depth completion (right). Droid-SLAM suffers from structural distortions and noise, especially in occluded or low-texture areas (red boxes), due to limited depth information. In contrast, our method generates more consistent reconstructions, preserving structural details and improving alignment (green boxes). These results highlight the advantage of dense depth completion in enhancing SLAM quality.

\section{Conclusion}
\label{sec:conclusion}

In this work, we present DMD$^{3}$C, a novel two-stage distillation framework that distills knowledge from monocular depth estimation foundation models into the depth completion task. In the first stage, we introduce a data generation strategy that leverages monocular depth estimation and mesh reconstruction to simulate training data, allowing the model to learn geometric features from diverse natural images. In the second stage, we propose a scale- and shift-invariant loss (SSI Loss) combined with a supervised L1 loss with sparse ground truth, which addresses the scale ambiguity in monocular depth estimation. This ensures consistent depth completions across varying scales and focuses on learning real-world scale information. Extensive experiments on the depth completion benchmarks demonstrate that DMD$^{3}$C achieves state-of-the-art performance, ranking first on the KITTI leaderboard, and significantly outperforming existing methods. Our results highlight the framework's ability to produce high-quality depth maps with improved detail and structural consistency, making it a promising solution for depth completion tasks.

\section{Acknowledgements}

This work was supported by the National Key R\&D Program of China (2022YFC3300704), the National Natural Science Foundation of China (62331006, 62171038, and 62088101), and the Fundamental Research Funds for the Central Universities.

{
    \small
    \bibliographystyle{ieeenat_fullname}
    \bibliography{main}
}

\end{document}